\pdfoutput=1

\documentclass[11pt]{article}
\usepackage[utf8]{inputenc}

\usepackage[final]{acl}

\usepackage{times}
\usepackage{latexsym}

\usepackage[T1]{fontenc}

\usepackage[utf8]{inputenc}

\usepackage{microtype}

\usepackage{inconsolata}

\usepackage{graphicx}

\usepackage{paralist}
\usepackage{inconsolata}
\usepackage{tabularx}
\usepackage{float}
\usepackage{caption}
\usepackage{graphicx}
\usepackage{multirow}
\usepackage{booktabs}
\usepackage{setspace}
\usepackage{booktabs}
\usepackage{hyperref}
\usepackage{paralist}
\usepackage{graphicx} 
\usepackage{amsmath}
\usepackage{colortbl}
\usepackage{booktabs}
\usepackage{multirow}
\usepackage{adjustbox}
\usepackage{array}
\usepackage{siunitx}
\usepackage{xcolor}  
\usepackage{listings}

\usepackage{arydshln}
\usepackage{caption}     
\usepackage{longtable}   
\usepackage{lscape}

\usepackage[table,xcdraw]{xcolor}

\usepackage[normalem]{ulem}
\useunder{\uline}{\ul}{}

\title{A Dynamic Fusion Model for Consistent Crisis Response}

\author{First Author \\
  Affiliation / Address line 1 \\
  Affiliation / Address line 2 \\
  Affiliation / Address line 3 \\
  \texttt{email@domain} \\\And
  Second Author \\
  Affiliation / Address line 1 \\
  Affiliation / Address line 2 \\
  Affiliation / Address line 3 \\
  \texttt{email@domain} \\}

\author{
  \textbf{Xiaoying Song\textsuperscript{1}}
  \textbf{Anirban Saha Anik\textsuperscript{1}}
  \textbf{Eduardo Blanco\textsuperscript{2}} \\
   \textbf{ Vanessa Frias-Martinez\textsuperscript{3}}
  \textbf{Lingzi Hong\textsuperscript{1}}
\\
  \textsuperscript{1} University of North Texas
\\
  \textsuperscript{2} University of Arizona
  \textsuperscript{3} University of Maryland
\\
  \small{
  \{xiaoyingsong, anirbansahaanik\}@my.unt.edu} \\ 
  \small{
  eduardoblanco@arizona.edu, vfrias@umd.edu, lingzi.hong@unt.edu  
  }
}

\begin{document}
\maketitle
\begin{abstract}

In response to the urgent need for effective communication with crisis-affected populations, automated responses driven by language models have been proposed to assist in crisis communications. 
A critical yet often overlooked factor is the consistency of response style, which could affect the trust of affected individuals in responders. 
Despite its importance, few studies have explored methods for maintaining stylistic consistency across generated responses. 
To address this gap, we propose a novel metric for evaluating style consistency and introduce a fusion-based generation approach grounded in this metric. 
Our method employs a two-stage process: it first assesses the style of candidate responses and then optimizes and integrates them at the instance level through a fusion process. 
This enables the generation of high-quality responses while significantly reducing stylistic variation between instances. Experimental results across multiple datasets demonstrate that our approach consistently outperforms baselines in both response quality and stylistic uniformity.

\end{abstract}

\section{Introduction}
People in crisis often turn to social networks for information, support, and assistance, especially when other sources cannot be relied upon
~\cite{bukar2022social}. 
Although some responses in social media from the general public offer valuable information and
emotional support, others may be inaccurate and even misleading to those in crisis~\cite{jafar2023social}. For example, during Hurricane Irma, users on Twitter (now X) shared conflicting information about whether shelters required identity checks, which affected whether some immigrants decided to evacuate~\cite{hunt2022monitoring}.

Direct communication from relevant government agencies or NGOs that carry out disaster relief efforts is critical to providing accurate information and verifying misleading information. 
However, authorities and NGOs often do not have enough resources to respond promptly to all affected individuals. At the same time, people’s needs are so different that a one-size-fits-all response is rarely effective~\cite{paulus2024interplay,lenz2023conceptualizing}. This challenge can be mitigated using LLM-based chat engines to understand natural conversations and generate informed responses~\cite{song2025speaking}. Leveraging AI to improve the efficiency, scalability, and accuracy of crisis communication has become a critical research focus~\cite{ziberi2024affect}.

\begin{figure}
    \centering
 
\includegraphics[width=0.9\linewidth]{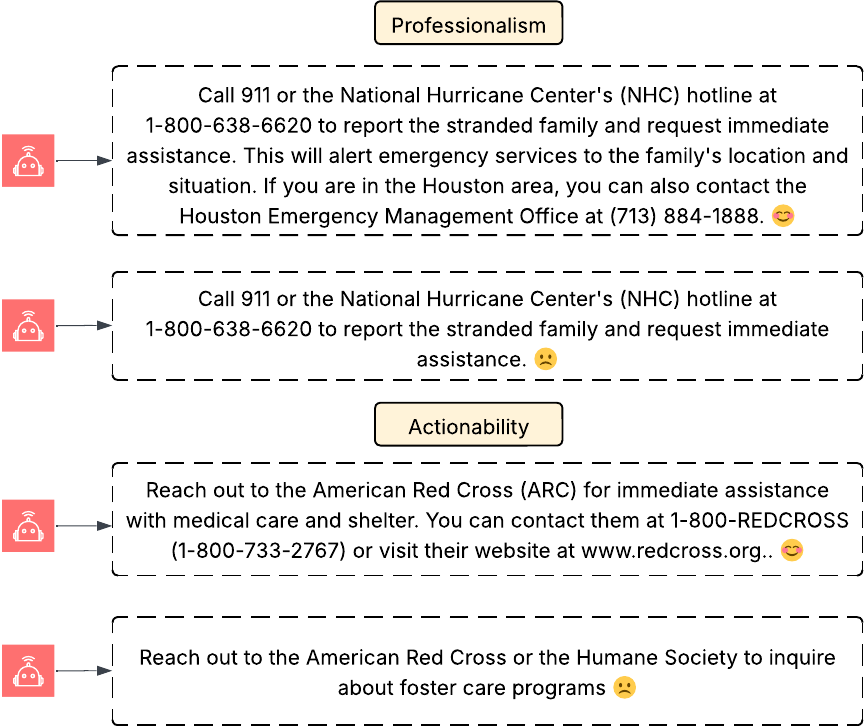}
    \caption{Examples of responses with high and low
    professionalism and actionability.
    Professional responses include explanations backing recommendations,
    demonstrating authority.
    Actionable responses offer specific guidance
    (e.g., phone numbers, website links) that users can follow to seek help.
    In this paper, we focus on generating \emph{consistent} responses,
    i.e., ensuring that professionalism, actionability, and relevance are roughly the same across all responses.
    }
    \label{fig:intro_example}
\end{figure}


Recent studies have explored the potential role of LLMs in supporting crisis communication~\cite{hong2025dynamic,xiao2025can,otal2024llm,grigorev2024incidentresponsegpt}. 
These systems aim to provide actionable, real-time guidance to affected individuals, focusing on user satisfaction, responsive interaction, and efficient use of resources~\cite{lei2025harnessing}.
However, an important issue remains overlooked: the consistency of automatically generated responses.

Authorities and NGOs have shown bias in their responses to people in crisis, which leads to inequitable access to aid and distrust~\cite{van2022humanitarian,huang2009determinants}.
We define consistency as the uniformity of the style in which information is conveyed across all responses. 
In particular, the core information conveyed should maintain the same level of quality regardless of the audience, crisis scenario, or communication platform.
Consistency signals organizational reliability. When messages remain aligned, audiences are more likely to trust the source~\cite{correia2024trust}. In contrast, inconsistent responses can be confusing and diminish trust~\cite{chatratichart2024inconsistency}.
For example, if some responses offer clear guidance while others are vague or off-topic, users may be uncertain about what to believe or do.
Figure \ref{fig:intro_example} shows examples of replies with different degrees of professionalism and actionability. 
When responses vary in quality across users, those receiving lower quality replies may perceive the interaction as inattentive or dismissive, resulting in dissatisfaction.


Previous studies have explored the generation of consistent responses in general-purpose dialogue systems, with particular attention to persona consistency \cite{lee2024dialogue}, semantic consistency \cite{fan2025consistency}, and factual consistency \cite{mesgar2021improving}.
Few studies have addressed style consistency in crisis communications~\cite{huang2009determinants}. Additionally, these studies typically employ fine-tuned generative models to increase consistency~\cite{lee2024dialogue, mesgar2021improving}.

There are no established metrics to evaluate the consistency of responses in crisis communication.
Effective crisis communication requires adherence to critical communicative functions~\cite{sellnow2021theorizing, coombs2007protecting}. 
These responses should be professional~\cite{steimle2024professional, coombs2007protecting}, actionable~\cite{coche2021actionability, bono2024effectiveness}, and relevant to user needs. 
Response consistency, therefore, entails delivering messages with stable characteristics across these dimensions, regardless of user query or scenario.
We propose a task-oriented definition for crisis communication: consistency refers to the degree to which all responses have similar characteristics across the three dimensions: professionalism, actionability, and relevance, while exhibiting minimal variation across responses. 

In addition, we propose a fusion framework to generate crisis responses with improved consistency. 
The approach integrates the strengths of the responses generated by multiple methods, taking advantage of their complementary advantages to produce highly effective outputs in all evaluation dimensions, resulting in reduced variations. 
Our approach employs state-of-the-art generation methods and explores various fusion methods. We evaluate the generation approaches in the three critical dimensions (professionalism, actionability, and relevance) as well as consistency across these dimensions. Experiments show the fusion framework enables the generation of responses with higher overall quality and consistency.
Specifically, we propose a novel fusion method grounded on assigning tailored weights to each dimension.
We experiment with Llama and Mistral and demonstrate that our fusion method results in superior performance compared to alternatives.

The contributions of this study include:
\begin{compactitem}
    \item We introduce a novel crisis response evaluation metric, \textbf{Consistency}, designed to ensure uniformity across key evaluation dimensions while addressing diverse information-need queries across crisis events.
    \item  We propose a \textbf{Fusion Framework}  that generates responses by integrating the strengths of outputs from different models, achieving strong performance on key evaluation metrics while ensuring consistency.
    \item We conduct detailed analyses demonstrating the fusion mechanisms obtains strong performance across LLMs, crisis scenario, and other realistic scenarios.
\end{compactitem}


\section{Related Work}

\paragraph{Information Needs and Responses in Crisis}
Individuals frequently use social media platforms to seek assistance in times of crisis.
Previous studies have proposed methods for detecting and classifying user needs.
Several datasets
offer granular categorizations of needs
~\cite{alam2021humaid,alam2021crisisbench}.
Recent studies have proposed using LLMs to facilitate timely responses ~\cite{hong2025dynamic, otal2024llm,yin2024crisissense,chowdhury2024infrastructure}. 
For example, \citet{goecks2023disasterresponsegpt} and \citet{otal2024llm} leveraged LLMs to generate actionable plans or guidance to crisis-affected individuals. 
\citet{grigorev2024incidentresponsegpt} developed IncidentResponseGPT, which leverages LLMs to automatically generate traffic incident response plans by synthesizing guidelines and processing real-time accident reports to inform authorities. 
\citet{rawat2024disasterqa} introduced DisasterQA, which is designed to evaluate LLMs in disaster response scenarios. They experimented with several prompting methods to answer crisis questions. 


These prior studies investigate approaches to generating responses for crisis communication.
We are the first, however, to 
investigate the consistency of responses, 
with a focus on maintaining a uniform style across varying scenarios.

\paragraph{Consistent Response Generation}
Consistent responses are essential for ensuring trust. In particular, it is important to avoid contradictions when addressing different audiences at different times, maintain a consistent tone, and ensure the conveyed information remains aligned~\cite{lee2024dialogue}.

Previous studies have explored various aspects of consistent response generation, including persona consistency, semantic consistency, and factual consistency.
Persona consistency refers to the alignment between generated responses and the established persona in dialogue systems \cite{lee2024dialogue,kim2023persona,mesgar2021improving}.
Semantic consistency ensures the generated responses logically follow the context without introducing irrelevant~\cite{fan2025consistency,song2025assessing}.
Factual consistency refers to the accuracy and correctness of generated content
\cite{mesgar2021improving}.
While these forms of consistency are crucial in general-purpose dialogue, they do not address consistency in balancing the critical communication dimensions required for crisis response, including professionalism, actionability, and relevance. To our knowledge, no prior work has systematically defined or evaluated consistency in the context of crisis communication, highlighting a gap that our work aims to address.

\begin{figure}
    \centering
    \includegraphics[width=1\linewidth]{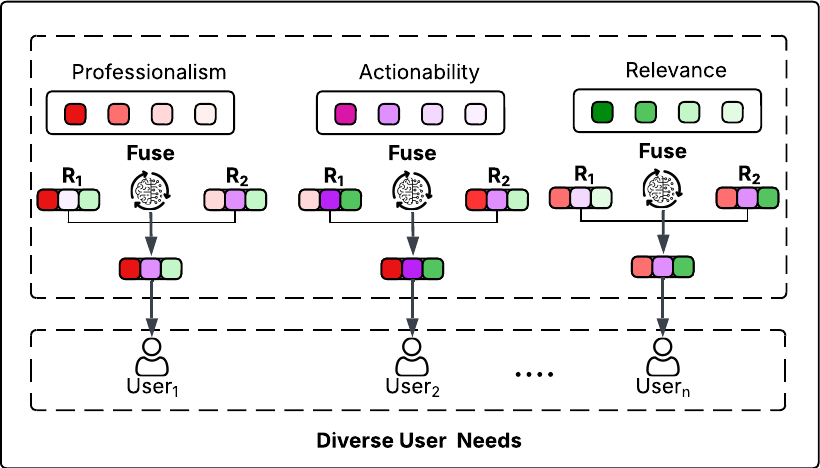}
    \caption{
    Overview of our fusion framework. Initial responses vary in
    professionalism (red), actionability (purple), and relevance (green); darker indicates higher.
    The fusion mechanism results in \textit{consistent} responses that address individual needs and combine the strengths of the initial responses: all users receive responses with high professionalism, actionability, and relevance.}
    \label{fig:fusion_diagram}
\end{figure}

\section{Consistency in Crisis Communication}

Consistency in crisis communication is crucial for maintaining trust and clarity. Our consistency involves producing professional, actionable, and relevant responses, as defined below.
Maintaining consistency across these dimensions is essential because variation can lead to confusion, reduced trust, and even harmful outcomes.

\begin{compactitem}
\item 
\textit{Professionalism} Professional responses ensure accurate, reliable, and credible assistance by leveraging knowledge and expertise to address crisis challenges effectively ~\cite{steimle2024professional,broekema2018role}. 

\item \textit{Actionability} Actionable responses deliver clear, practical, and relevant steps or guidance to address the concern or needs. In crisis response, solutions need to be straightforward and easy to implement~\cite{coche2021actionability}. 

\item \textit{Relevance} It evaluates how closely connected or appropriate generated responses are to the requests or queries showing needs. 

\end{compactitem}

Given a set of responses, the degree of variation is measured as the variance of scores in the three dimensions across all responses. 
\begin{equation}
\text{Variation} = \frac{1}{3} \left( \text{Var}_{\text{prof}} + \text{Var}_{\text{act}} + \text{Var}_{\text{rel}} \right)
\end{equation}

\noindent
where $\text{Var}_{\text{prof}}$, $\text{Var}_{\text{act}}$, and $\text{Var}_{\text{rel}}$ represent the variances in \textit{professionalism}, \textit{actionability}, and \textit{relevance}, respectively. 

The consistency score is defined as: 
\begin{equation}
\text{Consistency Score} = 1 - \text{Variation}
\end{equation}
Higher scores indicate better consistency, which refers to minimized fluctuation in standards that reliably address user needs across diverse queries and requests~\cite{kovavc2024stick}. 
It also supports scalability by ensuring that all users receive uniformly relevant, actionable, and professional guidance regardless of context or input variation.

\section{A Fusion Framework for Consistent Generation}
\label{Fusion Framework}
We propose a fusion framework to achieve consistent response generation in crisis communication. The framework is designed to integrate the strengths of conventional controllable response generation methods, balancing the key dimensions in crisis communication to achieve maximum consistency. Figure \ref{fig:fusion_diagram} illustrates the fusion framework. 

The framework leverages a fusion-based generation strategy that integrates generations from state-of-the-art approaches. Rather than selecting one output, we introduce a prompt-driven fusion mechanism that evaluates outputs by different models across critical communicative dimensions and synthesizes a new, improved response that draws on the strengths of both. We represent the process using the following formulation with the example of using Instructional Prompt (IP) and Retrieval-Augmented Generation (RAG):

\begin{align*}
\text{CC}(N, D) = \; 
\mathcal{L}\Big(&\text{Fuse}\big( M_{\text{IP}}(N), \; M_{\text{RAG}}(N), \\
& \mathbf{s}_{\text{IP}}, \; \mathbf{s}_{\text{RAG}} \big) \Big)
\end{align*}

\noindent
 $\text{CC}(N, D)$ represents the response generation process for a given crisis needs $N$ within a crisis-specific context $D$. The model generates two candidate responses: $M_{\text{IP}}(N)$ via Instructional Prompt, and $M_{\text{RAG}}(N)$ via Retrieval-Augmented Generation.
$\mathbf{s}_{\text{IP}}$ and $\mathbf{s}_{\text{RAG}}$ represent the score vectors of the Instructional Prompt and RAG outputs respectively, evaluated along three communicative dimensions: professionalism, actionability, and relevance.  $\text{Fuse}(\cdot)$ compares and balances the strengths of $M_{\text{IP}}(N)$ and $M_{\text{RAG}}(N)$ in these dimensions and generates a new response optimized across all aspects.
The process is further detailed in three steps. 

\noindent\textbf{Candidate Response Generation}
We employ state-of-the-art inference strategies to generate candidate responses, including the Instructional Prompt and RAG. 
These two methods are selected as they represent complementary approaches to response generation: Instructional Prompting leverages the reasoning and generalization capabilities of LLMs through carefully designed prompts, while RAG incorporates external evidence retrieved from a knowledge corpus to ground responses in factual content. This combination enables both flexibility and factuality, which are crucial for high-quality response generation. While other advanced methods exist, such as fine-tuned generation models or knowledge editing, we focus on Instructional Prompting and RAG due to their strong empirical performance, modularity, and ease of integration in diverse downstream tasks.

Instructional Prompt leverages zero-shot learning to generate crisis responses. 
As detailed in Appendix~\ref{Promts}, the prompt is crafted to define both the structure and intent of the response. The primary objective is to elicit outputs that consistently demonstrate high levels of professionalism, actionability, and relevance.
We experiment with variations of prompts and choose the one with the best performance in three evaluation dimensions for the following experiments
(See Appendix \ref{Appendix: Evaluation Details}).

Another method to generate candidate replies is RAG, which integrates external knowledge to provide factual information.  We refer to the authoritative resources from the Federal Emergency Management Agency (FEMA)\footnote{\url{https://www.fema.gov/}} to build our knowledge base, for example the \textit{Individual Assistance Program and Policy Guide,} which provides accessible programs and policies designed to support individuals during disaster.\footnote{\url{https://www.fema.gov/sites/default/files/documents/fema_iappg-1.1.pdf}}
FEMA's publications are grounded in government-endorsed emergency management protocols, ensuring their reliability as sources of factual information. They are tailored to various crisis scenarios, including hurricanes, wildfires, floods, and pandemics, offering relevant information for crisis responses. 

After collecting the knowledge, we construct a knowledge base for retrieval. Given the resources $S = \{D_1, D_2, \dots, D_N\}$ from FEMA, we split the content into individual documents to form the knowledge base $K = \{d_1, d_2, \dots, d_N\}$ for downstream retrieval. To enhance retrieval effectiveness, we adopt a hybrid approach that combines keyword-based and semantic retrieval methods, which has been shown to outperform single-method retrieval~\cite{anik2025multi, sawarkar2024blended}.
The hybrid retriever (\(R_h\)) integrates the strengths of keyword-based (\(R_k\) ) and semantic retrieval (\(R_s\)) via union: $R_h = R_k \cup R_s$.
 When retrieving the top-\( N \) documents (\( R_h = \{d_1, d_2, \dots, d_N\} \)), these documents are concatenated into a single context:
$C = \text{concat}(d_1, d_2, \dots, d_N)$.
The concatenated context \( C \) is then paired with the input query \( q \) to construct the prompt for the LLM to generate responses \(r\). We acknowledge we haven't incorporated real-time information, which could enhance adaptability in crisis communication, but this also incurs higher computational costs. We plan to explore the integration of real-time data in future work to further improve crisis communications.




\noindent\textbf{Multi-dimensional Evaluation}
\label{sec:evaluation}
After obtaining candidate responses, evaluations are conducted to provide criteria for fusion. 
For professionalism and actionability, the evaluation measures how users in crises would perceive these qualities. Given the lack of established automatic metrics for these dimensions and the high cost of recruiting real users, we utilize LLMs (GPT-4o mini\footnote{ Available at: \url{https://platform.openai.com}}) as evaluators to assist with the evaluations \cite{coche2021actionability}. The detailed instructions and generation are fed to LLMs to obtain the professionalism and actionability score.  
For relevance, we refer to previous studies to assess the similarity between generated responses and crisis needs using  BERTscore~\cite{zhangbertscore,zhou2024llm,liusie2024llm}. 
Additionally, we implement human evaluations to validate the assessment of LLMs. The details of evaluations are presented in Appendix \ref{Appendix: Evaluation Details}. 


\noindent\textbf{Fusion-based Generation}
The output of a single model may be unstable. To address this, we aggregate the outputs of multiple models, leveraging the strengths of each model. This fusion-based approach enables us to generate more balanced results across various critical dimensions, demonstrating higher overall quality and exhibiting consistency.

We design various in-context learning-based fusion methods. 
First, we experiment with \emph{Fusion with Evaluation Scores} (Fusion w/ Eval ). This method provides the LLM with numeric scores (e.g., professionalism, actionability, and relevance) associated with each candidate response. The model uses these scores as implicit guidance to identify and integrate the stronger elements of each response. However, without further instructions, the model may not consistently interpret or act upon the scores effectively.
Second, we design \emph{Fusion with Evaluation Scores and Structured Instructions} (Fusion w/ Eval \& Instruct ). Building upon the first method, this approach augments the score information with a prompt template that explicitly instructs the model to reason over the scores. The template directs the LLM to compare the candidate responses, retain the strengths from one, integrate key elements from the other, and synthesize them into a well-rounded output. This ensures more deliberate, interpretable fusion behavior and mitigates ambiguity in how the model uses the evaluation scores.
Third, we define \emph{Fusion with Weighted Evaluation Guidance} (Fusion w/ Eval \& Weight Instruct). Recognizing that optimizing all quality dimensions simultaneously may not always be feasible, we introduce weighted scores that reflect the relative importance of each dimension (e.g., 40\% professionalism, 40\% actionability, 20\% relevance). These weights guide the model to prioritize more critical dimensions during synthesis. This approach supports targeted optimization and helps enhance the overall response quality, especially in settings where trade-offs between dimensions are necessary.

\section{Experiments and Results}

\subsection{Dataset}


We use a Twitter  (now X)  dataset containing 1,013,313 geotagged posts from U.S. states affected by hurricanes Harvey, Irma, and Maria between August 15 and October 12, 2017. Geotagged tweets are used to ensure posts
are from crisis-affected individuals.

\begin{table*}[htbp]
\centering
\resizebox{\textwidth}{!}{
\begin{tabular}{lllccccc}
\toprule
\textbf{Model} & \textbf{Category} & \textbf{Method} & \textbf{Professionalism} & \textbf{Actionability} & \textbf{Relevance} & \textbf{Overall Quality} & \textbf{Consistency} \\
\midrule

\multirow{8}{*}{Llama}
& \multirow{5}{*}{Baseline} & Instructional Prompt  & 0.74 (0.33) & 0.52 (0.36) & 0.80 (0.02) & 0.66 & 0.76 \\
&                            & RAG & 0.96 (0.14) & 0.63 (0.33) & 0.80 (0.02) & 0.80 & 0.84 \\
&                            & RAG-PE & 0.94 (0.19) & 0.50 (0.14) & 0.80 (0.02) & 0.74 & 0.88 \\
&                            & Prompt and Select & 0.50 (0.50) & 0.98 (0.14) & 0.79 (0.02) & 0.75 & 0.78 \\
&                            & Fusion w/o Eval & 0.55 (0.27) & 0.97 (0.16) & 0.79 (0.02) & 0.77 & 0.85 \\
\cdashline{2-8}
&  \multirow{3}{*}{Fusion}   & Fusion w/ Eval & 0.98 (0.10) & 0.77 (0.27) & 0.79 (0.02) & 0.86 & 0.87 \\
&                            & Fusion w/ Eval \& Instruct & 0.92 (0.19) & 0.99 (0.07) & 0.79 (0.02) & 0.92 & 0.91 \\
&                            & \cellcolor{gray!15}Fusion w/ Eval \& Weight Instruct & \cellcolor{gray!15}\textbf{0.99 (0.07)} & \cellcolor{gray!15}\textbf{0.99 (0.09)} & \cellcolor{gray!15}\textbf{0.79 (0.02)} & \cellcolor{gray!15}\textbf{0.95} & \cellcolor{gray!15}\textbf{0.94} \\

\midrule
\multirow{8}{*}{Mistral}
& \multirow{5}{*}{Baseline} & Instructional Prompt  & 0.87 (0.34) & 0.98 (0.15) & 0.79 (0.02) & 0.90 & 0.83 \\
&                            & RAG & 0.87 (0.22) & 0.97 (0.11) & 0.81 (0.03) & 0.90 & 0.88 \\
&                            & RAG-PE & 0.76 (0.26) & 0.96 (0.15) & 0.80 (0.02) & 0.85 & 0.86 \\
&                            & Prompt and Select & 0.75 (0.39) & 0.81 (0.39) & 0.80 (0.03) & 0.78 & 0.73 \\
&                            & Fusion w/o Eval & 0.93 (0.25) & 1.00 (0.04) & 0.80 (0.02) & 0.93 & 0.90 \\
\cdashline{2-8}
&  \multirow{3}{*}{Fusion}   & Fusion w/ Eval & 0.92 (0.28) & 1.00 (0.08) & 0.80 (0.02) & 0.93 & 0.87 \\
&                            & Fusion w/ Eval \& Instruct & 0.96 (0.13) & 1.00 (0.05) & 0.80 (0.02) & 0.94 & \textbf{0.93} \\
&                            & \cellcolor{gray!15}Fusion w/ Eval \& Weight Instruct & \cellcolor{gray!15}\textbf{0.97 (0.13)} & \cellcolor{gray!15}\textbf{1.00 (0.08)} & \cellcolor{gray!15}\textbf{0.80 (0.02)} & \cellcolor{gray!15}\textbf{0.95} & \cellcolor{gray!15}0.92 \\

\bottomrule
\end{tabular}
}
\caption{Results (mean and standard deviation) using Llama and Mistral for response generation.
  Overall quality is the weighted average of professionalism, actionability, and relevance.
  While relevance remains roughly the same across all methods,
  our fusion approach generates the most consistent responses across the board
  while increasing both professionalism and actionability with Llama,
  and professionalism with Mistral.}
\label{main_performance}
\end{table*}

\noindent\textbf{Detect Information Needs Related Posts} 
We train three RoBERTa models to predict whether a tweet expresses information needs~\cite{alam2021crisisbench}.
Our classifiers are trained with three crisis datasets annotated with ``needs or request'' and other categories~\cite{ alam2021humaid,alam2021crisisbench}.
A tweet is labeled as ``needs-related'' if all three classifiers predict it as such. We opt for three smaller models rather than directly relying on LLMs for detection because they are more accurate and efficient.
We then conduct human validation to verify the predictions~\cite{song2025echoes} (See details in Appendix \ref{Appendix: Crisis Needs Detection Guidance}). Two research assistants are employed to annotate crisis needs. The agreement rate between two annotators is  94.5\%, with a Cohen's Kappa of 0.87. The agreement rate between classifiers and humans is 95\%, with a Kappa of 0.79, indicating the predictions are reliable.
We finally obtain 540 information needs related posts for experiments. 


\subsection{Experiment Setup}
We experiment with several open-sourced LLMs, including Llama-3.1-8B-Instruct \footnote{Available at \url{https://huggingface.co/meta-llama/Llama-3.1-8B-Instruct}}
 and Ministral-8B-Instruct-2410 \footnote{Available at \url{https://huggingface.co/mistralai/Ministral-8B-Instruct-2410}}
, which are good at conversational communications~\cite{taori2023alpaca,zheng2024balancing, li2024synthetic}.


\subsubsection{Baselines}
\noindent\textbf{Instructional Prompt}
We use the prompt detailed in Appendix \ref{Promts} as a baseline model and for generating candidate responses for fusion. We further experiment with various temperature settings and find out \textsc{Temperature 0.6} performs better in our task (Table \ref{tab:temperature-results}).

\noindent\textbf{RAG}
As mentioned in Section \ref{Fusion Framework}, we collect resources from FEMA to construct the knowledge base and use a hybrid search method incorporating 
two retrieval methods: keyword-based retrieval and semantic retrieval, using all-mpnet-base-v2\footnote{Available at \url{https://huggingface.co/sentence-transformers/all-mpnet-base-v2}} as the embedding model. 
In the generation process, we select the top-5 retrieved documents and concatenate them into a single context, providing additional knowledge for LLMs. The combined context and the full prompt are fed into the LLMs to generate responses.

\noindent\textbf{RAG with Prompt Engineering (RAG-PE)} 
 To examine whether the consistency and overall quality will be improved by prompt engineering and prove the necessity of the fusion work,
we experiment with RAG-PE, where the prompt is iteratively refined based on RAG's performance to generate effective responses across three dimensions. This method combines the strengths of RAG and Instruction Prompt with refined guidance. However, as RAG-PE relies on a single model, we hypothesize that RAG-PE may not achieve the same level of consistency as fusion models.

\noindent\textbf{Prompt and Select}
Following prior work on response generation~\cite{hong2024outcome, zhu2021generate}, we implement this approach, where LLMs are prompted to generate multiple candidates and the better response is chosen based on the evaluation scores. This method allows us to investigate whether selecting the most suitable response without fusion can improve consistency in the generated outputs. The fusion approaches allow for further optimization of candidate responses, presumably enabling the generation of outputs with better quality and reduced variances. 


\noindent\textbf{Fusion without Evaluation Score (Fusion w/o Eval)} 
Given that all our fusion methods incorporate evaluation scores as guidance, we design an experiment to examine whether LLMs can independently recognize the strengths without such kind of instructions. Therefore, we conduct an experiment where candidate responses are fused without referencing evaluation scores.

\subsubsection{Validation of Evaluators}
To validate the evaluations of professionalism and actionability by LLMs,
we engage human annotators to view the response and manually annotate based on the 3-scale definitions (See details in Appendix \ref{Appendix: Validation of Evaluators}). 
We randomly sample 100 tweets and their responses for annotations. The agreement rates between two annotations are above 85\% with Cohen's Kappa ($\kappa \geq 0.80$), indicating the human annotation is reliable. An expert assigns the final label for the human annotation, which will be used to compare with the LLM evaluator. The agreement rate and Cohen's Kappa ($\kappa \geq 0.72$) between human evaluation and LLM evaluation demonstrate substantial agreement.

\subsubsection{Model Settings}
We set all parameters the same for LLMs in the experiment. 
We set \texttt{max\_new\_tokens=256} for detailed yet concise responses. Sampling is enabled (\texttt{do\_sample=True}) with a temperature (\texttt{temperature=0.6}) as it generates the best results. The \texttt{top\_p=0.9} setting allows for some diversity while filtering unlikely tokens. Fusion prompts are detailed in Appendix \ref{Promts}.

\begin{table}[t]
\centering
\resizebox{0.4\textwidth}{!}{
\renewcommand{\arraystretch}{0.9}
\begin{tabular}{lccccc}
\toprule
\textbf{Setup} & \textbf{Pro} & \textbf{Act} & \textbf{Rel} & \textbf{Consist} \\
\midrule
\textsc{Temperature 0.4} & 0.41 & 0.40 & 0.02 & \textbf{0.72} \\
\textsc{Temperature 0.5} & 0.48 & 0.24 & 0.02 & \textbf{0.75} \\
\textsc{Temperature 0.6} & 0.33 & 0.36 & 0.02 & \textbf{\underline{0.76}} \\
\textsc{Temperature 0.7} & 0.30 & 0.39 & 0.02 & \textbf{\underline{0.76}} \\
\textsc{Temperature 0.8} & 0.38 & 0.43 & 0.02 & \textbf{0.72} \\
\bottomrule
\end{tabular}}
\caption{Professionalism, actionability, relevance and consistency using different temperatures experimenting on Instructional Prompt using Llama-3.1-8B-Instruct.} 
\label{tab:temperature-results}
\end{table}

\subsection{Results}
Table~\ref{main_performance} presents the results generated by the baseline and the fusion models.  

\noindent \textbf{Moderate temperatures yield the highest consistency in baseline generation.}
We first examine the effect of the temperature parameter on the consistency of generated responses. As shown in Table~\ref{tab:temperature-results}, setting the temperature to 0.6 or 0.7 produces the highest consistency scores (0.76). This suggests that moderate levels of randomness strike an effective balance between diversity and stability in generation. In contrast, lower temperatures (e.g., 0.4) constrain variation but slightly reduce consistency, while higher settings (e.g., 0.8) increase variability at the cost of stable response patterns. Overall, our findings indicate that a mid-range temperature optimizes consistency.


\noindent
\textbf{Fusion methods outperform all baselines in overall quality across models.} Fusion models retain similar relevance scores compared to baseline models; however, they can achieve much higher scores in professionalism and actionability, leading to high overall quality and low variance. 
In both Llama and Mistral, Fusion w/ Eval \& Weight Instruct achieves the best overall quality score of 0.95. 
This indicates that integrating the strengths of candidate responses produces higher-quality results than relying solely on a single model.

\begin{figure}
    \centering
    \includegraphics[width=1\linewidth]{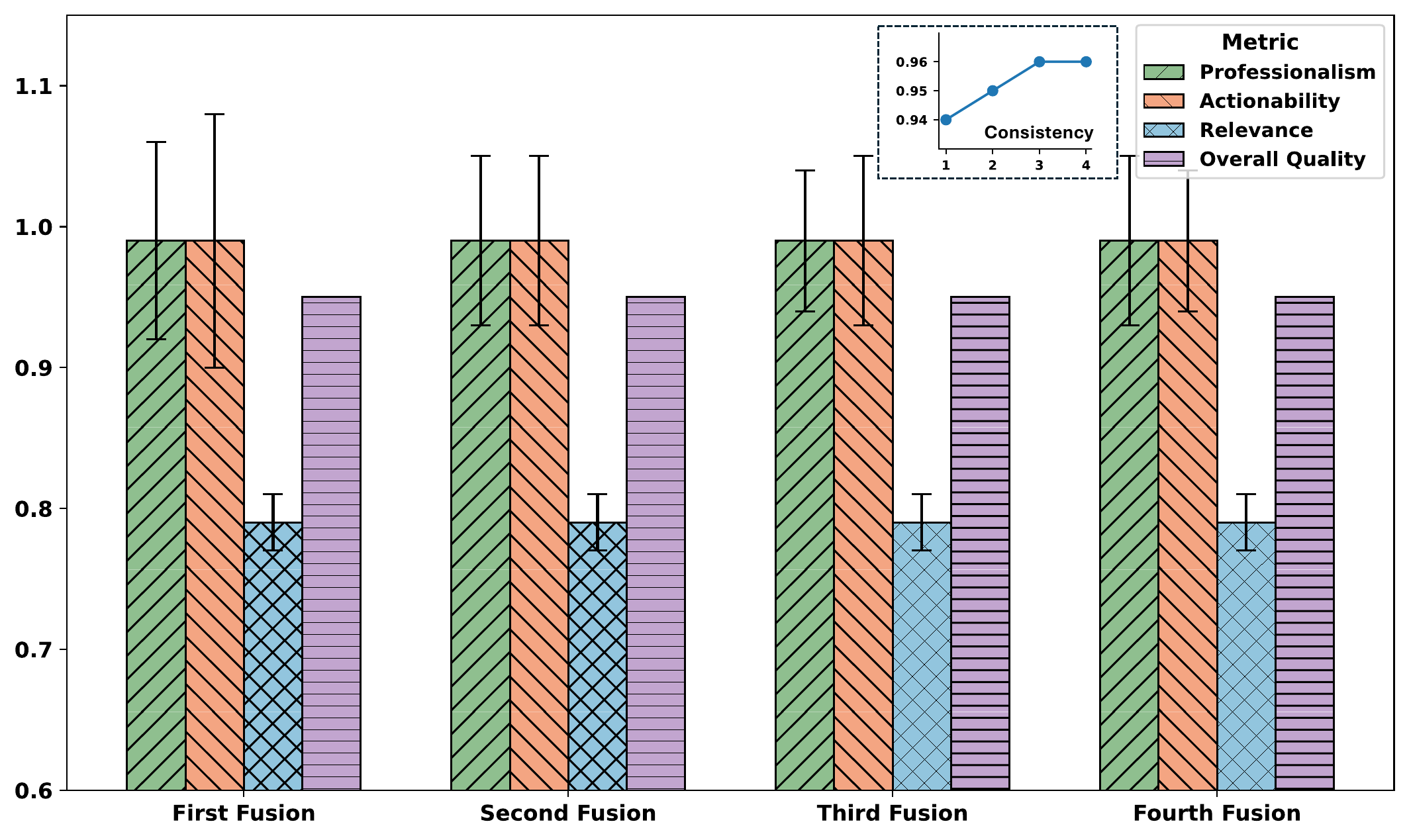}
    \caption{
    Results after one and more iterations of fusion with Eval \& Weight Instruct and
    Llama-3.1-8B-Instruct.
    Consistency scores are visualized in a mini line chart.
    Average professionalism, actionability, and relevance remain high from the first iteration.
    On the other hand, consistency plateaus after three iterations.
   }
    \label{fig:multiple iteration}
\end{figure}

\noindent
\textbf{Evaluation guidance is essential.}
Comparing fusion without evaluation guidance (Fusion w/o Eval), fusion with guidance (Fusion w/ Eval, Fusion w/ Eval \& Instruct, and Fusion w/ Eval \& Weight Instruct) achieves higher overall quality and consistency. 
The experiment confirms that fusion with evaluation guidance is more effective.

\noindent
\textbf{Consistency improves under structured fusion methods.} For both the Llama and Mistral model, Fusion w/ Eval \& Instruct and Fusion w/ Eval \& Weight Instruct demonstrate better consistency compared to all five baseline methods.
This indicates that LLMs with guided instructions are better at aggregating the strengths of individual responses, resulting in better consistency.

\noindent
\textbf{More fusion iterations do not lead to further improved performance.}
We further fuse the fused responses with responses generated by Instructional Prompt and RAG iteratively, using the Fusion with Eval \& Weight Instruct configuration as a representative example. As illustrated in Figure \ref{fig:multiple iteration}, performance in key dimensions, professionalism, actionability, and relevance, remains consistently stable in multiple iterations of the fusion, while consistency improves slightly and reaches an optimal after three iterations.

\subsection{Inconsistency Cause Analysis}
\label{Inconsistency Cause Analysis}
To further investigate the cause of inconsistency, we have conducted a finer-grained analysis by grouping the crisis requests into need categories defined by previous studies \cite{zguir2025detecting,yang2024detection}, and evaluating the variance of responses generated by Instructional Prompt using Llama-3.1-8B-Instruct on each category.
Additionally, we conduct additional analyses exploring how user query characteristics, such as detailedness, sentiment, and formality, affect the consistency of LLM-generated crisis responses.

Specifically, we categorize our crisis requests by need type and annotate each query for its level of detail (vague, medium, and specific), sentiment (neutral and emotional), and formality (casual and formal). We then calculate professionalism, actionability, relevance, and consistency scores for responses within each group. The results, shown in Appendix \ref{appendix: Inconsistency Cause Analysis} Table \ref{tab: inconsistency_analysis}, reveal several important trends:

\textbf{Response consistency is sensitive to linguistic variation within the same need type.}
For the Evacuation need, specific, neutral, and formal queries (Consistency: 0.90) outperform specific, emotional, and formal queries (Consistency: 0.74). This suggests that neutral sentiment in crisis scenarios may prompt more stable LLM behavior, potentially because emotional language introduces interpretive ambiguity or distracts from actionable content \cite{gandhi2025prompt, wang2025echoes}.

\begin{table}[t]
\centering
\renewcommand{\arraystretch}{1.1}
\resizebox{0.48\textwidth}{!}{
\begin{tabular}{lccccc}
\toprule
\textbf{Needs Category} & \textbf{Professionalism} & \textbf{Actionability} & \textbf{Relevance} & \textbf{Overall Quality} & \textbf{Consistency} \\
\midrule
Evacuation & 1.00 (0.00)  & 1.00 (0.00)  & 0.80 (0.02) & 0.96 & 0.99 \\
Food       & 1.00 (0.00)  & 1.00 (0.00)  & 0.80 (0.01) & 0.96 & 1.00 \\
Others     & 1.00 (0.00)  & 0.97 (0.18) & 0.81 (0.02) & 0.95 & 0.93 \\
Rescue     & 0.98 (0.14) & 0.98 (0.14) & 0.80 (0.02) & 0.94 & 0.90 \\
Shelter    & 1.00 (0.00)  & 1.00 (0.00)  & 0.80 (0.02) & 0.96 & 0.99 \\
\midrule
Average    & 0.99 (0.10) & 0.98 (0.14) & 0.80 (0.02) & 0.95 & 0.91 \\
\bottomrule
\end{tabular}}
\caption{Few-shot learning performance across various dimensions.}
\label{tab:fewshot-performance}
\end{table}

\textbf{The type of crisis need influences response variance.} For instance, Shelter queries that are specific and either neutral or formal achieve some of the highest consistency scores (0.82), while categories such as Rescue exhibit more moderate consistency and overall quality.

\textbf{The role of sentiment is context-dependent. }The sentiment dimension does not show a uniform impact across categories. In Rescue, both emotional and neutral sentiments yield comparable consistency (0.76 vs. 0.74), whereas in Food, emotional sentiment results in higher consistency (0.82) than neutral (0.77). This suggests that certain topics (like Food) benefit from emotional language, while others (like Evacuation) perform better with neutral expressions.

\begin{table*}[ht]
\centering
\renewcommand{\arraystretch}{0.9}
\resizebox{\textwidth}{!}{
\begin{tabular}{lccccc}
\toprule
\textbf{Method} & \textbf{Professionalism} & \textbf{Actionability} & \textbf{Relevance} & \textbf{Overall Quality} & \textbf{Consistency} \\
\midrule
\rowcolor{gray!15}
\multicolumn{6}{c}{\textbf{Baseline Methods}} \\
\addlinespace
Instructional Prompt   & 0.93 (0.24) & 0.94 (0.23) & 0.79 (0.02) & 0.91 & 0.84 \\
RAG                    & 0.94 (0.23) & 0.97 (0.12) & 0.77 (0.02) & 0.92 & 0.88 \\
RAG-PE                 & 0.76 (0.39) & 0.72 (0.40) & 0.77 (0.02) & 0.75 & 0.73 \\
Prompt and Select      & 0.97 (0.12) & 0.98 (0.12) & 0.77 (0.02) & 0.93 & 0.91 \\
Fusion w/o Eval         & 0.96 (0.21) & 0.97 (0.13) & 0.78 (0.02) & 0.93 & 0.88 \\
\midrule
\rowcolor{blue!5}
\multicolumn{6}{c}{\textbf{Fusion-Based Methods}} \\
\addlinespace
Fusion w/ Eval                  & 0.98 (0.10) & 0.98 (0.11) & 0.78 (0.02) & 0.94 & 0.92 \\
Fusion w/ Eval \& Instruct      & 0.96 (0.15) & 0.97 (0.15) & 0.78 (0.02) & 0.93 & 0.89 \\
Fusion w/ Eval \& Weight Instruct & \textbf{1.00 (0.00)} & \textbf{0.99 (0.11)} & 0.78 (0.02) & \textbf{0.95} & \textbf{0.96} \\
\bottomrule
\end{tabular}
}
\caption{Cross-crisis generalization results (earthquake and typhoon) with Llama-3.1-8B-Instruct. 
While relevance decreases compared to the same-crisis scenario (Table \ref{main_performance}), 
professionalism, actionability, and overall quality remain very high.}
\label{table:Generalization}
\end{table*}

Furthermore, previous researchers found that few-shot learning reduces variability in responses to the same sample despite prompt variations \cite{zhuo2024prosa}.  We have further conducted few-shot learning in our crisis response generation to investigate whether this method may improve the response consistency in crisis scenarios. We have drafted several response examples designed for diverse crisis needs and applied them in the few-shot learning experiment.

The results are shown in Table \ref{tab:fewshot-performance}. The average performance of the few-shot learning approach (Overall Quality: 0.95; Consistency: 0.91) remains slightly lower than our best-performing model (the Fusion w/ Eval \& Weight Instruct using Llama-3.1-8B-Instruct), which achieved an Overall Quality of 0.95 and a higher Consistency score of 0.94. Importantly, the fusion approach maintains robust performance and stability across a diverse range of user queries.

While few-shot learning effectively narrows the performance gap, especially when high-quality and targeted exemplars are available, our dynamic fusion model offers a more scalable and generalizable solution. It does not rely much on handcrafted prompts tailored to specific scenarios, making it more adaptable to real-world applications.

Moreover, our fusion method seamlessly integrates responses from RAG. Under this setting, we incorporate authoritative crisis-related knowledge from trusted sources such as FEMA, ensuring that the information provided is both accurate and contextually relevant. The inclusion of RAG also helps reduce hallucinations commonly produced by large language models, thereby further improving the factual reliability of responses.

\section{Cross Crisis Generalization}
To investigate the robustness of our fusion framework, we carry out experiments to generate responses to other crises such as earthquakes and typhoons. 
We employ the CrisisBench dataset ~\cite{alam2021crisisbench}, which comprises a diverse set of crisis events. 

We use the best-performing model, Llama-3.1-8B-Instruct (Consistency: 0.94, Overall Quality: 0.95), for the experiment.
Table \ref{table:Generalization} reports the performance of baseline and fusion methods. Among the baseline methods, Prompt and Select performs better in consistency (0.91) and overall quality (0.93).
Notably, fusion-based methods outperform the baseline methods. Especially, Fusion w/ Eval \& Weight Instruct achieves the best consistency (0.96) and overall quality (0.95). These findings indicate that our fusion framework not only performs well in hurricane-related contexts but also generalizes effectively to other crisis scenarios, confirming its applicability and robustness. We also repeat the experiments multiple times and present the results in Appendix \ref{Appendix: Multiple Rounds Fusion}, Figure~\ref{fig:multiple rounds}. The results show performance remains consistent across multiple rounds.

\section{Qualitative Analysis}
To investigate how humans perceive the generated crisis responses, we recruited two PhD students with a background in crisis computing to evaluate our responses. We select 50 responses generated by Instructional Prompt, RAG, and Fusion w/ Eval \& Weight Instruct using Llama-3.1-8B-Instruct.
(See evaluation guidance in Appendix \ref{Appendix: Qualitative Analysis})
We report the results in Table~\ref{tab:human_preference}, which indicate a higher preference for the fused responses, with an average rating of 0.86 and a consistency score of 0.86.




\begin{table}
\centering
\small
\renewcommand{\arraystretch}{0.5}
\resizebox{0.4\textwidth}{!}{
\begin{tabular}{lccc}
\toprule
\textbf{Metric} & \textbf{IP} & \textbf{RAG} & \textbf{Fusion} \\
\midrule
\textit{Agreement Metrics} &&& \\
\quad Agreement Rate     & 0.86 & 0.72 & 0.78 \\
\quad Cohen’s Kappa      & 0.76 & 0.60 & 0.62 \\
\addlinespace
\cdashline{1-4}
\addlinespace
\textit{Evaluation Results} &&& \\
\quad User Preference    & 0.48 & 0.47 & \textbf{0.86} \\
\quad Consistency        & 0.83 & 0.77 & \textbf{0.86} \\
\bottomrule
\end{tabular}
}
\caption{Human agreement and evaluation results across three strategies: IP = Instructional Prompt, RAG = Retrieval-Augmented Generation, Fusion = Fusion w Eval \& Weight Instruct. }
\label{tab:human_preference}
\end{table}

Through human evaluations, we observe distinct characteristics across the different strategies.
\textbf{Instructional Prompt:} Some responses offer clear and detailed instructions, 
while others are general and less actionable (e.g., \textit{``Stay safe and indoors, away from floodwaters and fallen power lines''}). In some cases, the model incorrectly refuses to generate a response, citing concerns about facilitating a scam, although the original crisis need was legitimate.
\textbf{RAG:} Some responses lack informativeness or appear evasive, using phrases such as \textit{"I don't know."} Although a few responses provide detailed action steps, but some are vague and lack actionable clarity (e.g., \textit{``Reach out to the American Red Cross or the Humane Society to inquire about foster care programs''}). 
\textbf{Fusion w/ Eval \& Weighted Instruct:} Most responses follow a consistent structure that includes both guidance and concise explanation. These responses provide concrete instructions with reliable references (e.g., \textit{``Reach out to the Harris County Emergency Management Office at (713) 755-5000 or the City of Houston's Emergency Management Office at (713) 837-0311 ...
}). Compared to other methods, the fusion approach generates responses with high quality consistently.

\section{Conclusion}
We introduce the evaluation of consistency for crisis communication, which requires that responses are uniformly professional, actionable, and relevant for all contexts. 
To achieve the generation of consistent responses, we propose a fusion framework and conduct experiments with various open-sourced LLMs. Results show that our fusion framework can achieve better consistency and higher overall quality across professionalism, actionability, and relevance. 
In particular, the evacuation scores are beneficial and enhance the fusion process. 
Cross-crisis experiments have been conducted to show the robustness of our framework across diverse crisis contexts. Human evaluation proves that our fusion-based generation obtains more preference.

\section*{Limitations}
\textbf{Limited Candidate Generation.} Even though we select the current state-of-the-art generation method to produce responses, there are still many other potential generation methods that could be used to further enhance the quality of fused responses.
We will explore more diverse models and leverage their strengths to facilitate candidate response generation.

\noindent
\textbf{Limited Resource for RAG response.} We collect information from FEMA, which is well-suited for our task. However, it is not sufficient to fully support crisis response generation due to the dynamic nature of real-world crises. In the future, we will collect more factual information from diverse sources and incorporate real-time information to assist crisis response generation.

\section*{Ethics Statement}

This study makes use of publicly available data collected from Twitter (now X). All data were accessed in accordance with Twitter’s Terms of Service and applicable platform policies. We ensured that the dataset does not contain personally identifying information beyond what is publicly visible, and we took steps to minimize potential risks to individual users. Specifically, any user identifiers were anonymized or removed, and only aggregated results are reported.
We acknowledge that Twitter data may contain offensive, biased, or otherwise harmful content. Such instances were carefully considered during data processing, and filtering strategies were applied where appropriate to reduce the propagation of harmful material. The use of this dataset is strictly for research purposes, and no attempts were made to deanonymize users or to use the data outside of its original research context.


\bibliography{custom}

\appendix




\section{Human Evaluation}
\subsection{Crisis Needs Detection Guidance}
\label{Appendix: Crisis Needs Detection Guidance}
We provide detailed guidelines in the following: Read the tweet and identify tweets where people seek help in crisis, such as food, medical supplies, and emotional support. Label the tweet as 1 if it demonstrates a need, and 0 if it does not. Examples are also provided to annotators for guidance. For instance, tweets like \textit{"We need tents, water, food, lanterns, medicine. In Peguy Ville..." or "My dog is hurt, is there any help around?..."} would be labeled as 1.

\subsection{Validation of Evaluators }
\label{Appendix: Validation of Evaluators}
We engage two PhD students with a background in crisis computing to serve as human annotators. Each is provided with crisis needs paired with corresponding responses. We define the evaluation criteria in Table \ref{tab:metric_criteria}.
\begin{table}[ht]
\small
\centering
\resizebox{0.48\textwidth}{!}{
\begin{tabular}{p{3cm}p{5.5cm}p{7.5cm}}
\toprule
\textbf{Metric} & \textbf{Definition} & \textbf{Criteria (Scoring Scale)} \\
\midrule
\textbf{Professionalism} & The extent to which the response conveys authority, credibility, and a well-substantiated foundation. &
\textbf{Score 0 (Not Professional):} The response is vague, lacks details, and does not mention specific organizations or actionable information. \newline
\textbf{Score 1 (Moderately Professional):} The response provides some professional elements but lacks specificity, such as mentioning general organizations without details on what they offer or how to contact them. \newline
\textbf{Score 2 (Highly Professional):} The response is well-structured, references specific organizations and programs, explains their relevance, and includes real contact information such as links, phone numbers, or emails. \\
\addlinespace
\textbf{Actionability} & The degree to which the response offers clear, practical, and relevant steps or guidance to address the concern or need expressed in the tweet. &
\textbf{Score 0 (Non-Actionable):} The response fails to provide any practical guidance or relevant steps. It may be vague, off-topic, or merely acknowledge the problem without offering a solution. \newline
\textbf{Score 1 (Partially Actionable):} The response provides some guidance but lacks clarity and specificity. It may contain useful information but is incomplete, unclear, or too general to be effectively acted upon. \newline
\textbf{Score 2 (Fully Actionable):} The response clearly and specifically provides detailed guidance or steps that the user can take immediately. It includes direct actions, useful resources, or concrete advice that fully addresses the concern. \\
\bottomrule
\end{tabular}}
\caption{Definitions and scoring criteria for response evaluation metrics.}
\label{tab:metric_criteria}
\end{table}

\subsection{Qualitative Analysis}
\label{Appendix: Qualitative Analysis}
We provide the following evaluation guidance: \textit{Assuming you are a user experiencing a crisis. Below is a crisis-related need and a generated response. Please rate the response on a scale from 1 to 5 based on your personal preference, considering the response’s professionalism, actionability, and relevance to the given need.} They independently evaluate the responses. The agreement rate and Cohen’s Kappa score were both above 0.60, indicating moderate inter-rater reliability.

To ensure a thorough understanding of the evaluation criteria, the annotators undergo training using example samples. Each annotator independently reviews and labels the data. Upon completion, a discussion is conducted to resolve disagreements. If consensus cannot be reached, an expert reviewer provides the final adjudicated label.

\section{Evaluation Details}
\label{Appendix: Evaluation Details}
We prompt LLM to evaluate the professionalism and actionability of generated responses. We design detailed guidelines for both professionalism and actionability, as outlined in the following prompts. Additionally, to validate the evaluations of professionalism and actionability by LLMs,
we engage human annotators to view the response and manually annotate based on the 3-scale definitions. We randomly sample 100 tweets and response pairs for annotations. The agreement rates between two annotations are above 85\% with Cohen's Kappa ($\kappa \geq 0.80$), indicating the human annotation is reliable. An expert assigns the final label for the human annotation, which will be used to compare with model evaluation.  The agreement rate and Cohen's Kappa ($\kappa \geq 0.72$) between human evaluation and model evaluation demonstrate substantial agreement.

\textit{1. Professionalism Evaluation:}
\lstdefinestyle{promptstyle}{
    basicstyle=\ttfamily\footnotesize,
    breaklines=true,
    frame=single,
    backgroundcolor=\color{gray!10},
    captionpos=b,
    numberstyle=\tiny\color{gray},
    keywordstyle=\color{blue},
    stringstyle=\color{red},
    showstringspaces=false
}

\begin{lstlisting}[style=promptstyle]
You are an expert evaluator tasked with assessing the professionalism of responses. Your goal is to determine how well each response adheres to professional communication standards, providing reliable and well-structured information.
Criteria:
Score 0 (Not Professional): The response is vague, lacks details, and does not mention specific organizations or actionable information.
Score 1 (Moderately Professional): The response provides some professional elements but lacks specificity, such as mentioning general organizations without details on what they offer or how to contact them.
Score 2 (Highly Professional): The response is well-structured, references specific organizations and programs, explains their relevance, and includes real contact information such as links, phone numbers, or emails.
Assign a score (0, 1, or 2) according to the criteria and reply with only the numerical score.
\end{lstlisting}


\textit{2. Actionability Evaluation:}
\lstdefinestyle{promptstyle}{
    basicstyle=\ttfamily\footnotesize,
    breaklines=true,
    frame=single,
    backgroundcolor=\color{gray!10},
    captionpos=b,
    numberstyle=\tiny\color{gray},
    keywordstyle=\color{blue},
    stringstyle=\color{red},
    showstringspaces=false
}

\begin{lstlisting}[style=promptstyle]
 You are an expert evaluator tasked with assessing the actionability of responses. Your goal is to determine how well each response provides clear, practical steps to address the concern or need expressed in the original tweet.
Criteria:
Score 0 (Non-Actionable): The response fails to provide any practical guidance or relevant steps. It may be vague, off-topic, or merely acknowledge the problem without offering a solution.
Score 1 (Partially Actionable): The response provides some guidance but lacks clarity and specificity. It may contain useful information but is incomplete, unclear, or too general to be effectively acted upon.
Score 2 (Fully Actionable): The response clearly and specifically provides detailed guidance or steps that the user can take immediately. It includes direct actions, useful resources, or concrete advice that fully addresses the concern.\\

Assign a score (0, 1, or 2) and provide a brief justification for the assigned score.
\end{lstlisting}

\begin{table*}[htbp]
\centering
\resizebox{0.9\textwidth}{!}{
\begin{tabular}{l l l l c c c c}
\toprule
\textbf{Need Category} & \textbf{Detailedness} & \textbf{Sentiment} & \textbf{Formality} & 
\multicolumn{4}{c}{\textbf{Evaluation Metrics (mean (sd))}} \\
\cmidrule(lr){5-8}
& & & & \textbf{Professionalism} & \textbf{Actionability} & \textbf{Relevance} & \textbf{Consistency} \\
\midrule
\multirow{4}{*}{Rescue} 
  & medium   & emotional & formal & 0.89 (0.22) & 0.39 (0.33) & 0.80 (0.02) & 0.81 \\
  & specific & emotional & casual & 0.77 (0.34) & 0.50 (0.45) & 0.80 (0.02) & 0.73 \\

  & specific & emotional & formal & 0.76 (0.35) & 0.53 (0.36) & 0.80 (0.02) & 0.76 \\
  & specific & neutral   & formal & 0.55 (0.44) & 0.65 (0.34) & 0.79 (0.02) & 0.74 \\
\cdashline{2-8}
\multirow{2}{*}{Shelter}
  & specific & emotional & formal & 0.83 (0.32) & 0.72 (0.33) & 0.80 (0.02) & 0.78 \\
  & specific & neutral   & formal & 0.67 (0.26) & 0.75 (0.27) & 0.79 (0.02) & 0.82 \\
\cdashline{2-8}
\multirow{2}{*}{Evacuation}
  & specific & emotional & formal & 0.68 (0.34) & 0.55 (0.42) & 0.81 (0.02) & 0.74 \\
  & specific & neutral   & formal & 0.50 (0.00) & 0.67 (0.29) & 0.79 (0.02) & 0.90 \\
\cdashline{2-8}
\multirow{2}{*}{Food}
  & specific & emotional & formal & 0.68 (0.28) & 0.42 (0.26) & 0.80 (0.02) & 0.82 \\
  & specific & neutral   & formal & 0.62 (0.23) & 0.44 (0.42) & 0.79 (0.03) & 0.77 \\
  \cdashline{2-8}

\multirow{5}{*}{Others}
  & medium   & emotional & formal & 0.62 (0.31) & 0.33 (0.33) & 0.80 (0.02) & 0.78 \\
  & specific & emotional & casual & 0.75 (0.42) & 0.35 (0.24) & 0.81 (0.03) & 0.77 \\
  & specific & emotional & formal & 0.78 (0.31) & 0.52 (0.37) & 0.80 (0.02) & 0.77 \\
  & specific & neutral   & formal & 0.71 (0.26) & 0.50 (0.39) & 0.81 (0.02) & 0.78 \\
  & vague    & emotional & formal & 1.00 (0.00) & 1.00 (0.00) & 0.80 (0.00) & 1.00 \\
\cdashline{2-8}
$ComConne^{\dagger}$  & specific & emotional & formal & 0.67 (0.41) & 0.58 (0.38) & 0.80 (0.04) & 0.73 \\
$EmoPsycho^{\dagger}$   & specific & emotional & formal & 1.00 (0.00) & 0.50 (0.00) & 0.80 (0.01) & 1.00 \\

$MisTrap^{\dagger}$  & specific & emotional & formal & 0.50 (0.71) & 0.00 (0.00) & 0.80 (0.01) & 0.76 \\
Medical Help & specific & emotional & formal & 0.57 (0.35) & 0.21 (0.27) & 0.79 (0.02) & 0.79 \\
\bottomrule
\end{tabular}}
\caption{The variance of response across the same crisis needs with diverse linguistic features. $ComConne^{\dagger}$ indicates Communication or Connectivity Issues. $EmoPsycho^{\dagger}$ means Emotional or Psychological Support. $MisTrap^{\dagger}$ refers to Missing or Trapped Persons.}
\label{tab: inconsistency_analysis}
\end{table*}

\section{Generation Prompts}
\label{Promts}


\noindent\textbf{Candidate Response Generation Prompts}
\begin{quote}
\texttt{
You are an AI assistant designed to provide professional, actionable, and relevant advice for someone seeking help related to a hurricane on social media.\\
Given the following tweet expressing needs during a hurricane, provide a detailed solution. If you don't know the answer, clearly state, 'I don't know'.\\
Guidelines:\\
- Prioritize immediate actions, clearly labeled as **Step 1**, **Step 2**, etc.\\
- For each action, provide a brief follow-up sentence to explain its importance or how to implement it.
- Include links, organizations, or contact information where relevant.\\
- Response should be professional, actionable, and relevant.
}
\end{quote}

\noindent\textbf{RAG-PE}
\begin{quote}
\texttt{
You are an AI assistant designed to provide practical, actionable, and relevant advice for individuals seeking help related to crisis on social media. Use the provided documents to address the needs expressed in the tweet. If you don't know the answer, clearly state, "I don't know."\\
Guidelines:\\
1. Prioritize Immediate Actions: Break down advice into clear, numbered steps labeled as Step 1, Step 2, etc.\\
2. Explain Each Action: For every step, include a brief follow-up sentence explaining its importance or how to implement it.\\
3. Provide Resources: Include links, organizations, or contact information where relevant to help the user take action.\\
4. Stay Concise: Keep responses clear and to the point, avoiding unnecessary details.
}
\end{quote}

\noindent\textbf{Prompt and Select}
\begin{quote}
\texttt{
You are an AI assistant designed to provide professional, actionable and relevant advice for someone seeking help during crises on social media. Two responses are provided, each with scores in three categories: Professionalism, Actionability, and Relevance.\\
Response 1: \{response1\} \\
Scores: \{scores1\} \\
Response 2: \{response2\} \\
Scores: \{scores2\} \\
Your task: Compare the two responses based on their scores. Return only the response that has the better overall performance.  
}
\end{quote}

\noindent\textbf{Fusion w/o Eval}
\begin{quote}
\texttt{
You are an AI assistant tasked with synthesizing two responses into one that optimally balances three key qualities: Professionalism, Actionability, and Relevance. Two responses are provided.\\
Response 1: \{response1\} \\
Response 2: \{response2\} \\
Your task is to merge these two responses into a single, cohesive answer. In doing so, you should maintain high levels of Professionalism, Actionability, and Relevance. Integrate the strongest elements from both responses and present the final response clearly.
Only provide the final response.
}
\end{quote}

\noindent\textbf{Fusion w Eval}
\begin{quote}
\texttt{
 You are an AI assistant tasked with synthesizing two responses into one that optimally balances three key qualities: Professionalism, Actionability, and Relevance.
 Two responses are provided, each with scores in three categories: Professionalism, Actionability, and Relevance.\\
Response 1: \{response1\} \\
Scores: \{scores1\} \\
Response 2: \{response2\} \\
Scores: \{scores2\} \\
Your tasks are:\\
 1. Internally analyze and compare the two responses based on their provided scores, identifying the strengths and essential elements of each.\\
2. Merge the strong qualities of Response 1 with the essential elements of Response 2 into a single, cohesive response that effectively balances Professionalism, Actionability, and Relevance.
Only provide the final response.
}
\end{quote}

\begin{figure}
    \centering
    \includegraphics[width=1\linewidth]{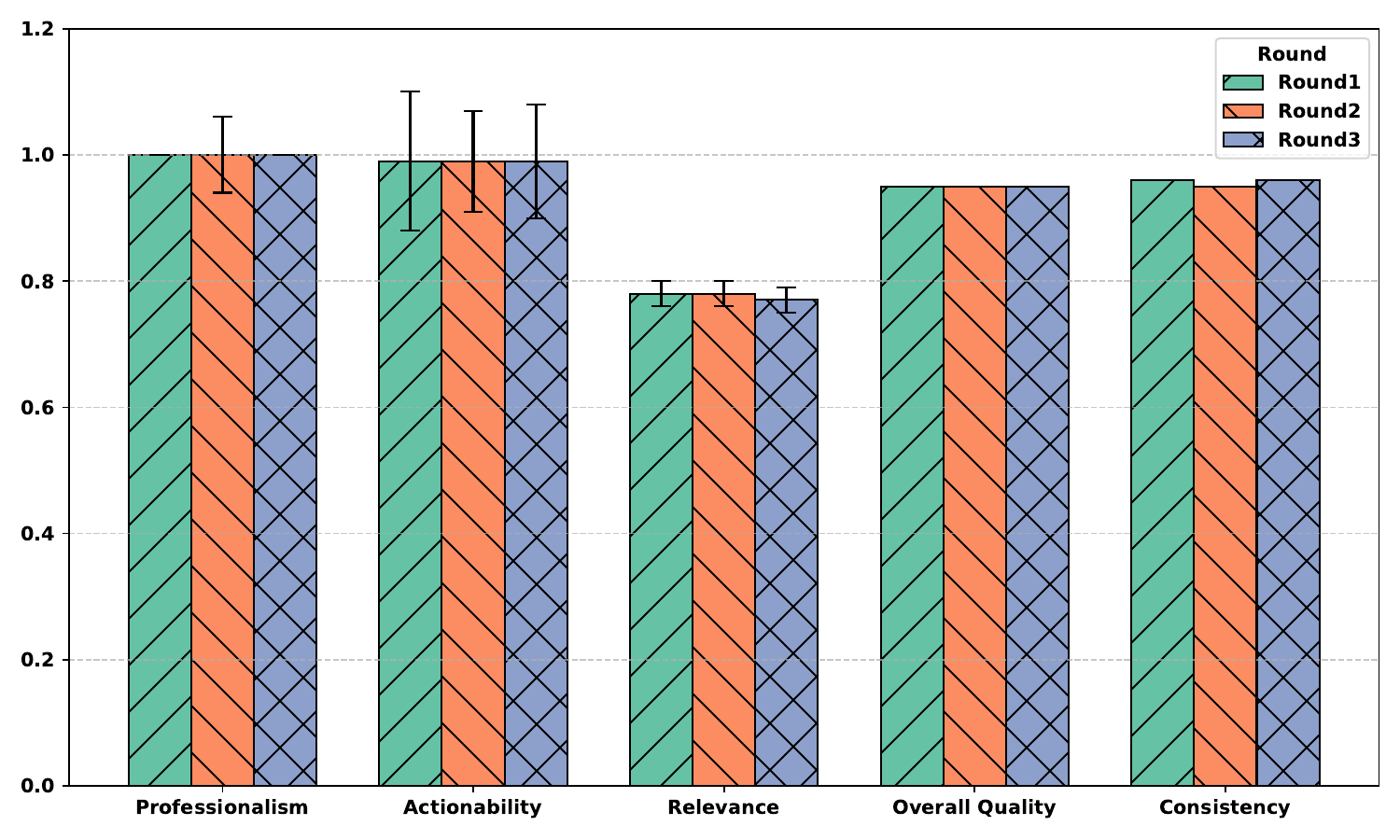}
    \caption{Multiple rounds of fusion w Eval \& Weight Instruct in generalization experiments. The results demonstrate that the method produces stable performance regardless of the number of fusion rounds.}
    \label{fig:multiple rounds}
\end{figure}

\noindent\textbf{Fusion w/ Eval \& Instruct}
\begin{quote}
\texttt{You are an AI assistant tasked with synthesizing two responses into one that optimally balances three key qualities: Professionalism, Actionability, and Relevance.
Two responses are provided, each with scores in three categories: Professionalism, Actionability, and Relevance.\\
Response 1: \{response1\} \\ 
Scores: \{scores1\}\\
Response 2: \{response2\}  \\
Scores: \{scores2\}\\
Your task:  
1. Compare the two responses based on their scores.  \\
2. Retain the \{\} and \{\} qualities from Response 1.  \\
3. Incorporate the \{\} and \{\} elements from Response 2.  \\
4. Merge these aspects into a single, well-rounded response that balances Professionalism, Actionability, and Relevance.  \\
5. Provide only the final merged response.}
\end{quote}

\noindent\textbf{Fusion w/ Eval \& Weight Instruct}
\begin{quote}
\texttt{You are an AI assistant evaluating and fusing two responses. 
Each response is accompanied by scores in four categories: Professionalism, Actionability, and Relevance.\\
Response 1:\\
{response1}\\
Scores: {scores1}\\
Response 2:\\
{response2}\\
Scores: {scores2}\\
Your task is:\\
1. Compare the two responses based on their scores in each category.\\
2. Synthesize the strengths of both responses to create a new, improved response that excels in all three areas.\\
3. The final quality of the improved response is determined by:\\
   - Professionalism: 40\% \\
   - Actionability: 40\% \\
   - Relevance: 20\% \\
4. Clearly list steps and explanations, resources, and provide contact information for the user to access help, the format:\\
    - Step 1: Explanation, resources, and contact information\\
    - Step 2: Explanation, resources, and contact information\\
Your objective is to produce a response that integrates the best elements of both responses, thereby achieving a higher overall quality.}
\end{quote}

\section{Inconsistency Cause Analysis}
\label{appendix: Inconsistency Cause Analysis}
In Section \ref{Inconsistency Cause Analysis}, we have conducted a finer-grained analysis by grouping the crisis requests into need categories and evaluating the variance of needs and responses.
Specifically, we categorized our crisis requests by need type and annotated each query for its level of detail (vague, medium, and specific), sentiment (neutral and emotional), and formality (casual and formal), referring to the linguistic analysis of \citet{perez2025analyzing}. We then calculated professionalism, actionability, relevance, and consistency scores for responses within each group. The results are shown in Table \ref{tab: inconsistency_analysis}.

\section{Multiple Rounds Fusion}
\label{Appendix: Multiple Rounds Fusion}
We run the fusion experiments multiple times to investigate whether the performance is stable. The results in Figure \ref{fig:multiple rounds} suggest that our fusion method, incorporating evaluation scores and weighted instructions, is robust and maintains stable performance across multiple rounds of application. This indicates that increasing the number of fusion rounds does not significantly degrade or improve performance; it remains consistently strong across key quality dimensions.

\section{Computing Resources}
The computational resources applied in this research include a high-performance server equipped with an Intel Xeon Gold 6226R processor, 128 GB memory, and 3 Nvidia RTX 8000 GPUs.

\section{Use of AI Assistants}
We acknowledge the use of AI tools to assist with code writing and expression refinement. 
The authors developed all core ideas, methods, analyses, and conclusions. 
The final content reflects the authors' independent scholarly contributions.

\end{document}